\pdfoutput=1
\documentclass[sigconf]{acmart}
\usepackage[nolist]{acronym}
\usepackage{booktabs}
\usepackage{tabu}
\usepackage{subfigure}
\usepackage{stfloats}

\renewcommand\footnotetextcopyrightpermission[1]{} 

\begin{document}

\begin{acronym}
\acro{ANN}{Artificial Neural Network}
\acro{ARIMA}{Auto-Regressive Integrated Moving Average}
\acro{CNN}{Convolutional Neural Network}
\acro{GRN}{Gated Residual Networks}
\acro{LSTM}{Long-Term Short-Term Memory}
\acro{ML}{Machine Learning}
\acro{MAE}{Mean Absolute Error}
\acro{MAPE}{Mean Absolute Percentage Error}
\acro{MSE}{Mean Square Error}
\acro{RQ}{Research Question}
\acro{RNN}{Recurrent Neural Network}
\acro{RMSE}{Root Mean Square Error}
\acro{SMAPE}{Symmetric Mean Absolute Percentage Error}
\acro{TFT}{Temporal Fusion Transformer}
\acro{VSN}{Variable Selection Network}
\end{acronym}

%\title{Comparing Day- and Week-Ahead Short-Term Load Forecasting with LSTM, TFT, and ARIMA Models}
%\title{Evaluation of the Temporal Fusion Transformer for Short-term Load Forecasting}
\title[Short-Term Electricity Load Forecasting Using the Temporal Fusion Transformer]{Short-Term Electricity Load Forecasting Using the Temporal Fusion Transformer: Effect of Grid Hierarchies and Data Sources}
\author{Elena Giacomazzi}
\affiliation{%
  \institution{University of Bamberg}
  %\department{Chair of Information Systems and Energy Efficient Systems}
  \city{Bamberg}
  \country{Germany}}
\email{elena@giacomazzi.de}
\orcid{0009-0006-6956-9977}

\author{Felix Haag}
\affiliation{%
  \institution{University of Bamberg}
  %\department{Chair of Information Systems and Energy Efficient Systems}
  \city{Bamberg}
  \country{Germany}}
\email{felix.haag@uni-bamberg.de}
\orcid{0009-0005-2227-2490}

\author{Konstantin Hopf}
\affiliation{%
  \institution{University of Bamberg}
  %\department{Chair of Information Systems and Energy Efficient Systems}
  \city{Bamberg}
  \country{Germany}}
\email{konstantin.hopf@uni-bamberg.de}
\orcid{0000-0002-5452-0672}

%\renewcommand{\shortauthors}{Giacomazzi et al.}

\begin{abstract}
  Recent developments related to the energy transition pose particular challenges for distribution grids. Hence, precise load forecasts become more and more important for effective grid management. Novel modeling approaches such as the Transformer architecture, in particular the Temporal Fusion Transformer (TFT), have emerged as promising methods for time series forecasting. To date, just a handful of studies apply TFTs to electricity load forecasting problems, mostly considering only single datasets and a few covariates. Therefore, we examine the potential of the TFT architecture for hourly short-term load forecasting across different time horizons (day-ahead and week-ahead)
  %, data sources (electricity consumption, calendar data, weather data, epidemic data), 
  and network levels (grid and substation level). We find that the TFT architecture does not offer higher predictive performance than a state-of-the-art LSTM model for day-ahead forecasting on the entire grid. However, the results display significant improvements for the TFT when applied at the substation level with a subsequent aggregation to the upper grid-level, resulting in a prediction error of 2.43\% (MAPE) for the best-performing scenario. In addition, the TFT appears to offer remarkable improvements over the LSTM approach for week-ahead forecasting (yielding a predictive error of 2.52\% (MAPE) at the lowest). We outline avenues for future research using the TFT approach for load forecasting, including the exploration of various grid levels (e.g., grid, substation, and household level).
  %, and the explainability of forecasting models.
\end{abstract}

%\texttt{\begin{CCSXML}
%<ccs2012>
%  <concept_id>10010520.10010553.10010562</concept_id>
%  <concept_desc>Computer systems organization~Embedded systems</concept_desc>
%  <concept_significance>500</concept_significance>
%  <concept_id>10010520.10010575.10010755</concept_id>
%  <concept_desc>Computer systems organization~Redundancy</concept_desc>
%  <concept_significance>300</concept_significance>
%  <concept_id>10010520.10010553.10010554</concept_id>
%  <concept_desc>Computer systems organization~Robotics</concept_desc>
%  <concept_significance>100</concept_significance>
%  <concept_id>10003033.10003083.10003095</concept_id>
%  <concept_desc>Networks~Network reliability</concept_desc>
%  <concept_significance>100</concept_significance>
%</ccs2012>
%\end{CCSXML}
%
%\ccsdesc[500]{Applied computing~Forecasting}
%\ccsdesc[300]{Computing methodologies~Neural networks}
%\ccsdesc{Hardware~Smart grid}
%\ccsdesc[100]{Hardware~Renewable energy}}


%  \Description{Enjoying the baseball game from the third-base seats. Ichiro Suzuki preparing to bat.}

\maketitle

\textit{\copyright Giacomazzi, Haag, Hopf (2023). This is the author's version of the work. It is posted here for your personal use. Not for redistribution. The definitive version was published in the Proceedings of the 14th ACM International Conference on Future Energy Systems (e-Energy '23), June 20--23, 2023, Orlando, FL, USA, \url{https://doi.org/10.1145/10.1145/3575813.3597345}.}

\section{Introduction}

%Relevanz der Stromverbrauchsvorhersage allgemein
Precise electricity load forecasts are key elements for planning and operating electrical power systems \cite{hong_energy_2020}. As utilities rely on such forecasts to purchase or generate electricity \cite{borkovski_electricity_2019}, the forecasting performance has a direct impact on their decision quality. With increasingly volatile electricity production and demand, electric load forecasting has recently become more difficult. Reasons include, among others, a more decentralized electricity generation, additional loads through sector coupling (heat pumps, electric vehicles, etc.) \cite{haben_short_2019}, changing behavioral patterns caused by the COVID-19 pandemic \cite{wen_impact_2022, zhong_implications_2020}, and the war in Ukraine.

%Relevanz der short-term Vorhersage
%\todo[inline]{Fokus auf Short-term -> was ist die Relevanz der short-term vorhersage, was sind die Herausforderungen (unerwartete Peaks, kosten Netzentgelte)}
%Electricity forecasts have been made for decades, and they are an active field of research \cite{hong_energy_2020, haben_short_2019}. Due to data availability and practical relevance, recent years have brought significant interest to the field of short-term load forecasting, i.e., day-ahead to week-ahead forecasts. % This is due to the growing need to manage the increasingly decentralized and volatile energy producers in distribution grids (e.g., through solar and wind power generation) on the one hand. On the other hand, new loads such as those generated by electric mobility or heat pumps emerge, while at the same time, the grid infrastructure often cannot expand at a sufficient pace.
Recent review studies on short-term electricity load forecasting \cite{haben_short_2019, hong_energy_2020, bourdeau_modeling_2019, hong_probabilistic_2016, vom_scheidt_data_2020} note that many research works examine the prediction of load at the level of complete energy systems (e.g., country or grid level). Another frequently investigated level is the household demand, given that smart meter data is increasingly available \cite{haben_short_2019, wang_review_2019, hopf_predictive_2019}. Yet, only a few studies have investigated intermediate levels in the low-voltage (distribution) grid for short-term load forecasting \cite{haben_short_2019}. Within this forecasting domain, hierarchical forecasting turns out to be a promising approach, as it can model the topological distribution of load across the grid \cite{hong_global_2014, athanasopoulos2009hierarchical}. %The literature distinguishes between a very short term horizon (few minutes to hours) , short term forecasts (one day to one week), medium term forecasts (the following month up to one year), finally long-term forecast (longer than one year) \cite{zhao_short-term_2021,dedinec_deep_2016,borkovski_electricity_2019,rahman_predicting_2018}. Forecasts on each horizon have different uses, bring their own challenges and different approaches are used thereby.

%\todo[inline]{Praktisches Problem: Vorhersagegüte noch nicht optimal, Vorhersage auf einer Netzebene, kaum auf substation + bisherige Verfahren können Zeitreihen verarbeiten und auch irrelevante information vergessen (LSTMs) aber sie können nicht mit langen Zeithorizonten umgehen}

%\todo[inline]{Bogen zu Transformer schlagen: weshalb sinnvoll, andere Bereiche sehr erfolgreich, erste Anwendungen im Bereich Energie aber ... keine Wetterdaten? Kalender-Events?}

While many approaches have been applied for short-term load forecasting, the most effective forecasting models currently employ variants of deep \ac{ANN} algorithms, such as \acp{LSTM} \cite{mujeeb_deep_2019,zou_weather_2019, zheng_short_2017, shohan_forecasting_2022}. \acp{LSTM} can handle sequences of data points well, yet, they have difficulties with learning patterns in long time series \cite{Bengio1994LearningLD}. To remedy this drawback, Vaswani et al. \cite{vaswani_attention_2017} presented a new architecture called Transformer, which outperforms \acp{LSTM} in several sequential modeling problems like natural language processing, text generation, and machine translation \citep{vaswani_attention_2017}. %Ultimately, the Transformer architecture enabled ChatGPT and similarly impressive applications.

Research has started to examine the Transformer approach for short-term electricity load forecasting \cite{zhao_short-term_2021} and also applied the \ac{TFT}, a Transformer variant for time series data, to short-term \cite{huy_short-term_2022} and mid-term \cite{li_probabilistic_2023} time horizons with promising results. However, the studies on Transformers and \acp{TFT} currently investigate selective aspects and focus on suggesting new algorithmic variants rather than thoroughly testing existing approaches for various problem facets. Particularly, the use of benchmark datasets and the examination of forecasts on various grid levels are missing, although both are known limitations in the field of energy forecasting \cite{hong_energy_2020}.

Our study addresses this research gap by conducting several experiments on the performance of the \ac{TFT} in hourly electricity forecasting on the distribution grid. We vary time horizons (day-ahead and week-ahead), data sources (electricity consumption, calendar data, weather data, epidemic data), and network levels (grid and substation level). Before we present our evaluation approach in \autoref{sec:evaluation} and analyze the results in \autoref{sec:results}, we review current time series forecasting methods and related works in the electricity forecasting field.

\section{Background}

Starting with the first studies on short-term load forecasting in the 1960s \cite{heinemann_and_1966, zhao_short-term_2021}, scholars have conducted intensive research within this field. Such research includes conventional statistical approaches (e.g., linear regression and \ac{ARIMA} models), but also \ac{ML} methods such as fuzzy logic \citep{khan2015load}, and random forest \citep{borkovski_electricity_2019}. 
In recent years, similar to other applications of forecasting, (deep) \ac{ANN} approaches have gained prominence in the field of electricity load forecasting \citep{borkovski_electricity_2019, hong_energy_2020}. Particularly \acp{LSTM} have proven to be a robust forecasting approach in several variations, as shown by studies based on data obtained from Scotland \cite{zou_weather_2019}, Malaysia \cite{farsi_short-term_2021}, the U.S. \cite{zheng_short_2017}, and Great Britain, France, and Germany \cite{aasim_data_2021, farsi_short-term_2021}.
%
%This work will focus on deep learning methods because the literature shows that those outperform the other methods in almost all tasks \citep{borkovski_electricity_2019}. 

Recent studies examined the Transformer architecture \cite{vaswani_attention_2017} for load forecasting. The \ac{TFT} \cite{lim_temporal_2020} in particular holds significant potential to boost the predictive performance, as it 
%
%\paragraph{Research opportunity and focus of this study}
%
%The \ac{TFT} seems to be a promising approach for time series forecasting and holds the potential to 
overcomes known limitations of both, the Transformer and the \ac{LSTM} architecture for time series forecasting. For the application field of short-term load forecasting, we found\footnote{We searched Google Scholar, ACM digital library, and IEEE Xplore using the keywords “short-term electricity load forecasting” and “transformer”.} several recent studies (see \autoref{tab:related-work}) that evaluate Transformers and \acp{TFT} (and variants of them) with diverse sets of parameters and data. All of these studies indicate that the Transformer and \ac{TFT} approaches outperform other methods for short-term load forecasting. However, we identify three issues that need further investigation.

\begin{table}
    \footnotesize
    \caption{Studies examining Transformer architectures for short-term electricity load forecasting}\label{tab:related-work}
    \begin{tabu}{lXlllllllll}
    \toprule
Ref.	&  
        Model	& Place	& Years	& N	& Cal.	& Temp.	&  Level	\\
        \midrule
\cite{lim_temporal_2021} &   
        TFT	                    & PT & 2011-14 & 	1	& &	&Grid \\
\cite{zhang_power_2021} &   
        Transf.+k-Means	    & FR & 2006-10	& 1 & &&Household \\
\cite{zhao_short-term_2021} &  
        Transf.+k-Means	    & AU & 2006-10 & 1 & &&Grid \\	
\cite{li_heat_2022} &  
        Transf.-variant     & CN & - & 1 & && Heat appl.\\
\cite{huy_short-term_2022}	& 
        TFT+lin. reg.   & VN & 2014-21 & 	1	& x &	x		&Grid\\	
\cite{lheureux_transformer-based_2022} & 
        Transf.            & US & 2004-08 &	20 & &	x		 & Substation 	\\
\cite{huang_short-term_2022}    &  
        Transf.-variant	    & SP & 2016	&1&				&&Grid	\\
 \cite{zhang_short-term_2022} 	& 
        Transf.-variant	    & AU & 2006-10	& 1 &	x&	x&		Grid	\\
%\cite{li_total_2022} & 2022 & Intl. Conf. on ITQM & 
%        Transformer-variants    & Country          & CN & 2009-2020 & 1 (monthly) &
\cite{cao_probabilistic_2022}   & 
        Transf.             & PA & 2017-20 & 1 & x & x &  Grid\\
\cite{li_probabilistic_2023} 	&  
        Transf., TFT, ITFT	& CN	& 2016-17&	1&	x&	x&		Grid	\\
				&&UK&	2004-09&	1&	x&	x\\			
    \midrule
    This &  \ac{TFT} & US & 2004-08 &	20 & x &	x		&  Substation\\
    study & \ac{TFT} & DE & 2019-22 & 70 & x & x &  Substation\\
    %&&&TFT & UCI & PT & 2011-2014 & 370 & && Grid\\
    
    \bottomrule
    \end{tabu}
\end{table}

First, almost all studies that we found propose a slightly different version of the Transformer or \ac{TFT} architecture and test them with a single dataset (only one study \cite{li_probabilistic_2023} uses a second, publicly available dataset to demonstrate external validity). Hence, it remains unclear to what extent the reported performance results are generalizable or dataset-specific (an aspect that is also criticized in several review studies on short-term load forecasting \cite{haben_short_2019, hong_energy_2020}). A comparison across different datasets would be helpful, although this requires significant effort and computational resources.

Second, several studies that we identified are quite selective regarding the input variables they consider. Some use electricity load data only \cite{lim_temporal_2021, zhang_power_2021, zhao_short-term_2021, li_heat_2022, huang_short-term_2022}, although the inclusion of exogenous variables such as weather or calendar data are known to improve forecast quality and should therefore be included \cite{jain_forecasting_2014, haben_short_2019}. A detailed analysis of the performance of the \ac{TFT} with different exogenous variables would benefit the assessment of the architecture's potential for load forecasting.

Third, the majority of studies focus on a single forecasting unit, e.g., the load of the whole grid \cite{lim_temporal_2021, zhao_short-term_2021, huy_short-term_2022, huang_short-term_2022, zhang_short-term_2022, li_probabilistic_2023, cao_probabilistic_2022}, a single household \cite{zhang_power_2021} or a heating system \cite{li_heat_2022}. This observation is also echoed in comprehensive review studies on short-term load forecasting \cite{haben_short_2019}, and energy forecasting \cite{hong_energy_2020}. Yet, forecasting on secondary levels of the distribution grid (e.g., substations or grid zones) is beneficial for grid operation and planning and has the potential to boost predictive performance, as hierarchical forecasts can model the topological distribution of load across the grid.

Our study addresses the outlined research gaps by using the \ac{TFT} architecture for short-term load forecasting (day-ahead and week-ahead) on the grid and substation level while considering an acknowledged benchmark dataset. 
%Thereby, we seek to answer the following \acp{RQ}:

%\begin{description}
%    \textbf{RQ1:} \textit{How well does the \ac{TFT} predict day-ahead and week-ahead electricity load in the distribution grid (i) using a single time series on the first grid-level or (ii) using multiple time series from the secondary grid-level?}
    
%    \textbf{RQ2:} \textit{To what extent does the forecasting error of the \ac{TFT} change with the availability of different input variables, i.e., (i) electric load and calendar data, (ii) electric load, calendar and weather data, (iii) electric load, calendar, weather, and epidemic data?}
%\end{description}

\section{Forecasting method for short-term load forecasting}
\label{sec:method}

\begin{figure}[h]
    \centering
    \includegraphics[width=\linewidth]{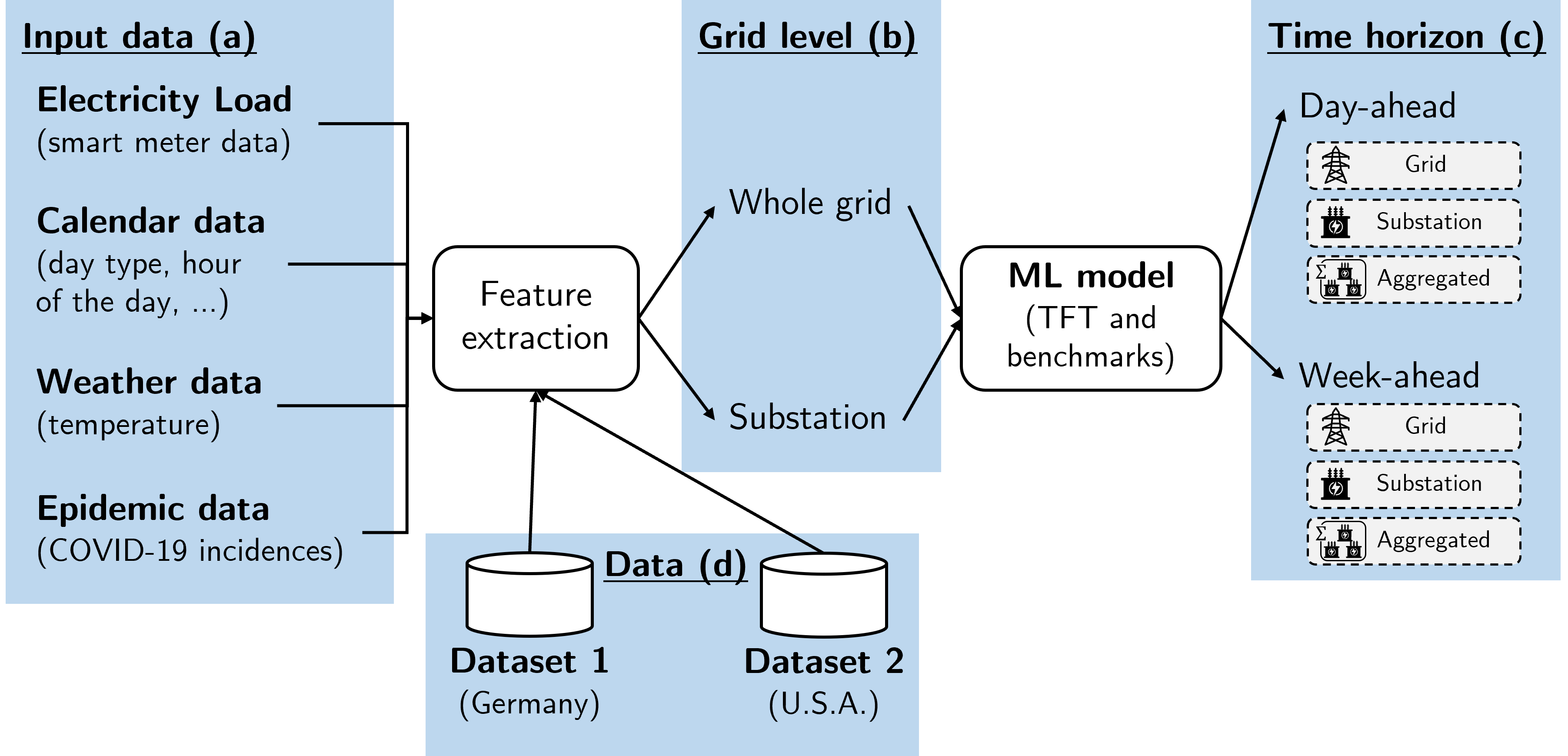}
    \caption{Simplified illustration of the approach; numbers indicate evaluation variations}
    \label{fig:flowchart}
\end{figure}

We instantiate the \ac{TFT} architecture to forecast the hourly electricity load based on past electricity consumption data and a variety of further data sources, as we illustrate in \autoref{fig:flowchart}. Thereby, our particular focus lies on the variation of the grid-level of the forecast (b). More specifically, we examine the \ac{TFT} performance in the distribution grid using a single time series on the first grid-level and using the aggregation of multiple time series from the secondary grid-level (substation-level).
%In our first \ac{RQ}, we examine how well the \ac{TFT} can predict electric load in the distribution grid, (i) using a single time series on the first grid-level or (ii) using multiple time series from the secondary grid-level. 
%
The latter makes use of the \ac{TFT}'s and \ac{LSTM}'s capability to forecast several target variables for future time steps at the same time (also known as multi-horizon time series forecasting).

For comparison, we also examine the predictive performance of \ac{LSTM} and ARIMA models. By varying four elements of the forecasting setup, we analyze how well the \ac{TFT} architecture performs under certain configurations. Thereby, we vary the input data (a) the time horizon (c) and test day-ahead and week-ahead forecasts, as well as two datasets (d).

%\begin{figure}[h]
%    \centering
%    \includegraphics[width=\linewidth]{img/Substation and Network level.png.png}
%    \caption{Aggregation Levels for Training}
%    \label{fig:net_subst_overview}
%\end{figure}

\paragraph{Feature extraction}
To obtain a multidimensional forecast, we consider four different data sources: Electric load, calendar, weather, and epidemic data. \autoref{tab:features} provides an overview of all features employed, their value ranges, and their use in our study. %We list all features together with the datasets' value ranges in \autoref{tab:data_features}.
We list further details on the data preparation in appendix \ref{app:data_feats}.

\begin{table}[bp]
    \centering
    \footnotesize
    \caption{Features with value ranges and horizon}
    \label{tab:features}
    %\begin{minipage}{.9\linewidth}
    \begin{tabular}{lllcrr}
        \toprule
        Data & Feature & Range  & Horizon$^a$ & Mean \textit{DE} & Mean \textit{US}\\
        \midrule
        Load & Consumption &  $\mathbb{R}$ &P&&\\
        &\hspace*{1em}(grid)                  &&& 3{,}175.82 & 392{,}945.48\\
        &\hspace*{1em}(substation)            &&& 46.93 & 20{,}681.34\\
        \hline
        Calendar&Hour of the day & $[0,23]$ &F& &  \\
        &Day of the week & $[0,6]$&F&  & \\
        &Day of the year & $[1,365]$&F& & \\
        &Holiday/weekend & $[0,1]$&F& 0.31& 0.31\\
        \hline
        Weather&Temperature & $\mathbb{R}$&F& 10.48 & 14.46\\
        \hline
        Epidemic&Covid-19 incidence & $\mathbb{R}$ &P& 6& x\\
        \hline
        Other &Grid node ID$^b$ & $\mathbb{I}$ & S&&\\
        \bottomrule
        \multicolumn{6}{l}{\footnotesize{\textit{a) F: Feature known in the future, P: Feature known up to the present, S: Static feature}}}\\
        \multicolumn{3}{l}{\footnotesize{\textit{b) only on substation-level prediction}}}
    \end{tabular}
    %\end{minipage}
\end{table}

\paragraph{ANN model architectures}

Our forecasting approach provides load forecasts on an hourly level (i.e., the next 24 hours \textit{day-ahead} and the next 168 hours \textit{week-ahead}). All features are connected with the \ac{TFT} using a separate \ac{VSN} per input type. Weights are shared among \ac{VSN} for past known, future known, and static variables, respectively. \autoref{tab:features} lists these variable types. For benchmarking the predictive performance of the \ac{TFT}, we use a linear \ac{ARIMA} estimator and an \ac{LSTM} architecture. Our analysis uses the Python package \textit{darts} \citep{herzen_darts_2022}.

\paragraph{Grid-level forecast}

We vary the levels for which we obtain forecasts. First, by considering the complete grid, which is a single time series of demand data. Second, by obtaining a substation-level forecast, which considers multiple time series for training and forecasts in each time step to predict demand data for each substation. To obtain a more precise forecast on the grid-level, we aggregate all substation-level forecasts---an approach that the literature describes as hierarchical load forecasting \cite{hong_global_2014}.

\section{Performance Evaluation}
\label{sec:evaluation}
%\input{tables/substation_statistis}

%\paragraph{Datasets}
We rely on two datasets to evaluate the performance of the \ac{TFT}-based forecasting approach. The first stems from a local grid operator located in central Germany ($DE$) and covers a recent time frame (2019--2021). The second is a validation dataset, which is publicly available and stems from the Global Energy Forecasting Competition 2012 (GEFC'12) \cite{hong_global_2014} ($US$). It comprises data from 20 grid zones in the U.S., which we consider as substations. The detailed processing of both datasets is described in appendix \ref{app:data_feats}.

%\paragraph{Model training and evaluation}
For model training, we choose a time-wise 80/20 train-test split. For the $DE$ dataset, the training set spans from Jan 1st, 2019, to May 23rd, 2021. The test set comprises the period from May 24th, 2021, until Dec 31st, 2021. For the day-ahead forecast, we choose all complete days (0h-23h), and for the week-ahead forecast, all complete weeks (Mo-So) in the test set. In total, we rely on 219 days and 28 weeks for the evaluation using the $DE$ test set data.
The training set of the $US$ dataset spans from Jan 1st 2004 to March 14th 2007. The test set consists of the remaining data until December 31st, 2007. For the evaluation of the $US$ dataset, the day-ahead test set consists of 291 complete days, and the week-ahead evaluation of 38 full weeks. We normalize all input features for both datasets to ensure unbiased model training \cite{normalization}.

We performed a random hyperparameter search \cite{bergstra2012random} for those parameters for which we could not obtain meaningful values through reasoning. For the \ac{TFT}, the parameters are: Number of neurons in the hidden layer, the number of \ac{LSTM} layers, the number of attention heads, the dropout value, the batch size, and the size of the input window. For this purpose, we conducted a hyperparameter search using sweeps from the "Weights \& Biases" platform \cite{wandb}. The configurations for each sweep are based on the parameter bandwidth suggested in \cite{lim_temporal_2021}. In addition, we varied the input window size $k$ across the day-ahead forecast with $k \in [24, 48, 72, 168, 336, 672]$ and for the week-ahead forecast with $k \in [168, 336, 504, 672]$. We list the final parameter configurations of the best-performing models for each task in appendix (\autoref{tab:parameters}).

%\subsection{Evaluation Criteria}

To evaluate the \ac{TFT}, \ac{LSTM}, and \ac{ARIMA} models, we compare the predicted values $p_t$ with the actual demand values $y_t$ for each time step $t=1,\ldots, N$ and assess the forecasting performance using the \ac{RMSE} as absolute and \ac{MAPE} and \ac{SMAPE} as relative error metrics, which find regular use in earlier studies. 

\section{Results}
\label{sec:results}

Within the scope of this paper, we present and discuss the main results of our analysis. Thereby, we analyze the predictive performance of the \ac{TFT} against ac{ARIMA} and \ac{LSTM} and compare the performance of our approach with earlier studies. %Then, we focus on answering the two \acp{RQ} that we set out as the focus of our study.
A detailed list of the performance results of our approach can be found in \autoref{tab:error-results} as part of the appendix. 

\subsection{Baseline comparison}

% For baseline assessment, we analyze the predictive performance of the \ac{TFT} against ac{ARIMA} and \ac{LSTM}, and compare the performance of our approach with earlier studies. %\footnote{Only one study \cite{huy_short-term_2022} publishes relative error metrics for point estimates that allow a comparison.}. 
For the analysis of the models, we focus on \ac{SMAPE} as it allows for a comparison across datasets and grid-levels while expressing the relative performance results in relation to the actual and forecasted value.
Similar to earlier studies, the ARIMA models display a relatively high error, which can be attributed to their limited capacity to generalize over long time series. Consequently, the \ac{LSTM} and \ac{TFT} clearly outperform the (more simplistic) statistical approach.
%\todo[inline]{The ARIMA error metrics are particularly high in our study, as we fed the complete training time series into the ARIMA model and did not tune the parameters.}

Overall, we obtain lower errors for the \ac{TFT} than the \ac{LSTM} models, yet, not for all configurations.
%, as we explain below. 
We observe a clear superiority of the \ac{TFT} with a larger forecasting horizon (week-ahead). We attribute this result to the stronger capability of the \ac{TFT} architecture to learn patterns over longer time intervals. The \ac{TFT} also performs better than the \ac{LSTM} for a forecast on the substation level. For day-ahead forecasts and single time series forecasts, the \ac{LSTM} approach is still a reasonable alternative.

%As a second analysis, we compare the performance of our comprehensive DE dataset with that using a benchmark dataset (PT) to enable the comparison with other studies. Thus, this analysis just uses consumption data, as the PT dataset has no geographic reference available. 
With demand and calendar features (configuration II in \autoref{tab:error-results}), the \ac{TFT} has an average performance of 3.98 MAPE, which is similar to what \cite{huy_short-term_2022} and \cite{zhang_short-term_2022} report in their studies, although we have a simpler data processing without applying linear regression to the input features to estimate trends. As these studies do not provide pure \ac{TFT} and \ac{LSTM} estimates and only partially use public datasets, an appropriate comparison is not feasible. Lim et al. \cite{lim_temporal_2021}, who propose the \ac{TFT} approach, conclude that the \ac{TFT} results in a lower error than other approaches for time series forecasting. Using \ac{RMSE}, \ac{MAPE}, and \ac{SMAPE} to review the forecasting error, we cannot confirm this result for the day-ahead, but for the week-ahead forecast. Other works applying the \ac{TFT} to electricity forecasting do not report relative error metrics, which makes a reasonable comparison of the results unfeasible.

\subsection{Forecasts on various grid-levels}

%We examine the \ac{TFT} performance in the distribution grid using a single time series on the first grid-level and using the aggregation of multiple time series from the secondary grid-level (substation-level).
%In our first \ac{RQ}, we examine how well the \ac{TFT} can predict electric load in the distribution grid, (i) using a single time series on the first grid-level or (ii) using multiple time series from the secondary grid-level. 
%
%For the substation-level forecast (and the hierarchical, which aggregates the load of all substations), we make use of the capability of the \ac{TFT} to forecast several target variables for future time steps at the same time (also known as multi-horizon time series forecasting).

%H1: Einseitiger t-test \ac{SMAPE} \ac{TFT} hierarchical vs. grid-level mit beiden Datensätzen und day/week (d.h. 4 tests)
The results show that the \ac{TFT}'s hierarchical forecast outperforms single time series forecasting regarding predictive performance on the grid-level: We observe a statistically significant difference for both approaches regarding the day-ahead \textit{t}(363.58)~=~13.90, \textit{p}~<~.001, \textit{d}~=~1.41 (with \ac{MAPE} $2.43\%$) and week-ahead \textit{t}(38.27)~=~6.56, \textit{p}~<~.001, \textit{d}~=~1.82 (with \ac{MAPE} $2.52\%$) forecasts using the DE dataset. We validate this result by performing the same tests on the US dataset, where we also find a significant difference with a large effect size for both approaches (day-ahead \textit{t}(550.99)~=~10.60, \textit{p}~<~.001, \textit{d}~=~0.89, and week-ahead \textit{t}(71.14)~=~3.84, \textit{p}~<~.001, \textit{d}~=~0.88). Hence, we conclude that the hierarchical forecast approach outperforms single time series forecasting on the grid-level.
Additionally, the results display performance improvements for the \ac{LSTM} architecture when the load is predicted and aggregated at the substation level---however, this observation does not hold for all predictive cases of this study.

\section{Discussion}

\paragraph{Practical contribution}
Our analysis demonstrated the potential of the \ac{TFT} approach compared to a state-of-the-art approach (\ac{LSTM}) and a simple estimator (ARIMA). In addition, we empirically illustrated the benefits that can be obtained through hierarchical load forecasting in the electric grid (reflecting the call for such analyses from recent review studies \cite{hong_energy_2020, haben_short_2019}).

While we observed that the \ac{TFT} approach is on par with (or only slightly better than) the LSTM approach on a day-ahead horizon, the TFT clearly outperformed the LSTM on a week-ahead horizon. Conversely, the TFT seems to be more costly to train regarding the computational effort because it has significantly more parameters. Therefore, practitioners need to balance a trade-off between more accurate methods in longer time spans and computational costs.

\paragraph{Limitations and future work}
Our study is a starting point for a more in-depth evaluation of Transformer and TFT approaches in the domain of load forecasting. In summary, we identify six areas for future research:
%
%First, we applied the in-development Python package \textit{darts} \cite{herzen2021darts} for our implementation. The library did not allow taking into account the epidemic data for the \ac{LSTM} models, which should be addressed in  future work. To estimate the contribution of epidemic data, a comparison with these data on the \ac{LSTM} architecture is thus still pending.

First, we used weather observations as inputs for the forecasting period, which leads to an underestimation of the forecasting error \cite{haben_short_2019}. In practice, only weather forecasts are available. Future studies should therefore include historical weather forecasts and quantify their impact on the models' forecasting quality.

Second, we included the Covid-19 incidence as a covariate for the TFT. However, the incidence data do not properly represent the lockdown periods. Hence, additional epidemic data might reflect time periods and their effect on the energy demand more precisely (e.g., by employing a binary feature that reflects lockdown periods). %Future work can also consider an \ac{LSTM} implementation that includes both, past and future covariates. %The consideration of an \ac{LSTM} approach could lead to new insights on the effect of epidemic features for short-term load forecasting.
%urthermore, we could not test the epidemic data with the \ac{LSTM} to back our findings, given that the used darts package did only support the use of \qq{future covariates} (i.e., time series used for training also need to be present for the forecasting horizon).
% Also, other features from open data sources should be included in future studies \cite{hopf_predictive_2019}.

Third, we only compared point estimates of the forecasting models in our study. However, probabilistic forecasting is a very promising area in load forecasting \cite{haben_short_2019, hong_probabilistic_2016}. Future studies may extend the \ac{TFT} approach and assess its potential for probabilistic forecasting.

Fourth, for real-world applications, the runtime performance of the models and their explainability might be of major importance to electricity vendors. 
%our evaluation of the \ac{TFT} approach focuses on the predictive and runtime performance of the algorithms applied. There are other aspects of higher relevance for practice, such as the explainability of models that are \qq{black box} (which usually applies to \ac{ANN} based approaches). 
In some cases, higher explainability outweighs higher costs for training \citep{borkovski_electricity_2019, hong_energy_2020, lago_forecasting_2018}. The \ac{TFT} architecture contains an interpretable multi-head self-attention mechanism that enables feature importance-based explanations \cite{lim_temporal_2021}. So far, this functionality has not been studied for the case at hand, although explainable \ac{ML} offers detailed insights on model forecasts that can benefit decision-makers \cite{haag_augmented_2022}.

Fifth, our analysis has shown the potential of predictions on more granular network levels employing a subsequent aggregation. 
Future work should make use of increasingly available smart meter data to obtain household level predictions and their aggregations to enhance the forecasting performance.

Sixth, we integrated empirical load data mostly as is in our analysis. The body of forecasting literature has suggested several meaningful data preprocessing steps that improved the performance of less complex forecasting models, such as taking into account typical daily or weekly load profiles \cite{hoverstad_short-term_2015}. Considering that varying existing algorithms often result in only small changes in predictive performance, we encourage future research to focus on an in-depth evaluation of existing methods, more advanced feature engineering, and the evaluation of real-world problems with (multiple) benchmark datasets.

% \begin{figure}
%     \centering
%     \begin{subfigure}{\linewidth}
%          \centering
%          \includegraphics[width=0.48\linewidth]{img/Loss_Day_Single_NC.png}  
%          %\caption{Loss on Grid Day Single NC}
%          \label{fig:loss_day}
%     \end{subfigure}
%     \begin{subfigure}{\linewidth}
%          \centering
%          \includegraphics[width=0.48\linewidth]{img/Loss_Week_Single_NC.png}  
%          %\caption{Loss on Grid Week Single NC}
%          \label{fig:loss_week}
%     \end{subfigure}
%     \caption{Loss functions}
%     \label{fig:loss}
% \end{figure}
\section{Conclusion}
Current developments related to more volatile electricity production and demand challenge the management of the electric grid. Thus, precise load forecasts become more and more important. Recent forecasting literature has proposed the \ac{TFT} architecture, which theoretically addresses known limitations of the \ac{LSTM} and Transformer approach. To date, studies on the \ac{TFT} approach to short-term load forecasting have been empirically inconclusive and neglect external validity. 

Our study carries out several experiments using the \ac{TFT} architecture and multiple datasets. The results show that the \ac{TFT} architecture does not outperform a \ac{LSTM} model for day-ahead forecasting for the entire grid. Yet, we find that the predictive performance of the \ac{TFT} is higher when applied at the substation level in conjunction with a subsequent aggregation to upper grid-levels.

Our investigation opens avenues for future research on the \ac{TFT} approach for short-term load forecasting. In particular, we would like to motivate other scholars to conduct further experiments, specifically with respect to different network levels of forecasting (e.g., grid, substation, household) and the explainability of the models used.

% Thus, a \ac{MAPE} of up to 2.43 (SMAPE 2.44) could be achieved in our data set for the day-ahead forecast. On a week-ahead basis, the \ac{TFT} approach clearly outperforms \ac{LSTM} and achieved a \ac{MAPE} of up to 2.5 (SMAPE 2.51).

\begin{acks}
%\grantsponsor{⟨sponsorID⟩}{⟨name⟩}{⟨url⟩}
%\grantnum[⟨url⟩]{⟨sponsorID⟩}{⟨number⟩}
We gratefully thank our research partner Stadtwerk Haßfurt GmbH for providing comprehensive data from their distribution grid that enabled this study. We further thank the \grantsponsor{1}{Bavarian Ministry of Economic Affairs, Regional Development, and Energy}{https://www.iuk-bayern.de/} for their financial support of the project "DigiSWM" (\grantnum{1}{DIK-2103-0014}).
\end{acks}

\bibliographystyle{ACM-Reference-Format}
\bibliography{Zotero_Elena, MA_extra}


\begin{thebibliography}{43}

%%% the Web address.

\ifx \showCODEN    \undefined \def \showCODEN     #1{\unskip}     \fi
\ifx \showDOI      \undefined \def \showDOI       #1{#1}\fi
\ifx \showISBNx    \undefined \def \showISBNx     #1{\unskip}     \fi
\ifx \showISBNxiii \undefined \def \showISBNxiii  #1{\unskip}     \fi
\ifx \showISSN     \undefined \def \showISSN      #1{\unskip}     \fi
\ifx \showLCCN     \undefined \def \showLCCN      #1{\unskip}     \fi
\ifx \shownote     \undefined \def \shownote      #1{#1}          \fi
\ifx \showarticletitle \undefined \def \showarticletitle #1{#1}   \fi
\ifx \showURL      \undefined \def \showURL       {\relax}        \fi
% invisible to TeX
\providecommand\bibfield[2]{#2}
\providecommand\bibinfo[2]{#2}
\providecommand\natexlab[1]{#1}
\providecommand\showeprint[2][]{arXiv:#2}

\bibitem[{Aasim} et~al\mbox{.}(2021)]%
        {aasim_data_2021}
\bibfield{author}{\bibinfo{person}{{Aasim}}, \bibinfo{person}{S.~N. Singh},
  {and} \bibinfo{person}{Abheejeet Mohapatra}.}
  \bibinfo{year}{2021}\natexlab{}.
\newblock \showarticletitle{Data driven day-ahead electrical load forecasting
  through repeated wavelet transform assisted {SVM} model}.
\newblock \bibinfo{journal}{\emph{Applied Soft Computing}}
  \bibinfo{volume}{111} (\bibinfo{year}{2021}), \bibinfo{pages}{107730}.
\newblock
\showISSN{1568-4946}
\urldef\tempurl%
\url{https://doi.org/10.1016/j.asoc.2021.107730}
\showDOI{\tempurl}


\bibitem[Athanasopoulos et~al\mbox{.}(2009)]%
        {athanasopoulos2009hierarchical}
\bibfield{author}{\bibinfo{person}{George Athanasopoulos},
  \bibinfo{person}{Roman~A Ahmed}, {and} \bibinfo{person}{Rob~J Hyndman}.}
  \bibinfo{year}{2009}\natexlab{}.
\newblock \showarticletitle{Hierarchical forecasts for Australian domestic
  tourism}.
\newblock \bibinfo{journal}{\emph{International Journal of Forecasting}}
  \bibinfo{volume}{25}, \bibinfo{number}{1} (\bibinfo{year}{2009}),
  \bibinfo{pages}{146--166}.
\newblock


\bibitem[Bengio et~al\mbox{.}(1994)]%
        {Bengio1994LearningLD}
\bibfield{author}{\bibinfo{person}{Yoshua Bengio}, \bibinfo{person}{Patrice~Y.
  Simard}, {and} \bibinfo{person}{Paolo Frasconi}.}
  \bibinfo{year}{1994}\natexlab{}.
\newblock \showarticletitle{Learning long-term dependencies with gradient
  descent is difficult}.
\newblock \bibinfo{journal}{\emph{IEEE transactions on neural networks}}
  \bibinfo{volume}{5 2} (\bibinfo{year}{1994}), \bibinfo{pages}{157--66}.
\newblock


\bibitem[Bergstra and Bengio(2012)]%
        {bergstra2012random}
\bibfield{author}{\bibinfo{person}{James Bergstra} {and}
  \bibinfo{person}{Yoshua Bengio}.} \bibinfo{year}{2012}\natexlab{}.
\newblock \showarticletitle{Random search for hyper-parameter optimization}.
\newblock \bibinfo{journal}{\emph{Journal of machine learning research}}
  \bibinfo{volume}{13}, \bibinfo{number}{2} (\bibinfo{year}{2012}).
\newblock


\bibitem[Biewald(2020)]%
        {wandb}
\bibfield{author}{\bibinfo{person}{Lukas Biewald}.}
  \bibinfo{year}{2020}\natexlab{}.
\newblock \bibinfo{title}{Experiment Tracking with Weights and Biases}.
\newblock
\newblock
\urldef\tempurl%
\url{https://www.wandb.com/}
\showURL{%
\tempurl}
\newblock
\shownote{Software available from wandb.com}.


\bibitem[Borkovski et~al\mbox{.}(2019)]%
        {borkovski_electricity_2019}
\bibfield{author}{\bibinfo{person}{Stefan Borkovski}, \bibinfo{person}{Stefan
  Petkoski}, {and} \bibinfo{person}{Maja Erkechova}.}
  \bibinfo{year}{2019}\natexlab{}.
\newblock \showarticletitle{Electricity consumption forecasting using recurrent
  neural network: {Electrical} trade market study}.
\newblock \bibinfo{journal}{\emph{Innovations}} (\bibinfo{year}{2019}),
  \bibinfo{pages}{8}.
\newblock


\bibitem[Bourdeau et~al\mbox{.}(2019)]%
        {bourdeau_modeling_2019}
\bibfield{author}{\bibinfo{person}{Mathieu Bourdeau},
  \bibinfo{person}{Xiao~qiang Zhai}, \bibinfo{person}{Elyes Nefzaoui},
  \bibinfo{person}{Xiaofeng Guo}, {and} \bibinfo{person}{Patrice Chatellier}.}
  \bibinfo{year}{2019}\natexlab{}.
\newblock \showarticletitle{Modeling and forecasting building energy
  consumption: {A} review of data-driven techniques}.
\newblock \bibinfo{journal}{\emph{Sustainable Cities and Society}}
  \bibinfo{volume}{48} (\bibinfo{date}{July} \bibinfo{year}{2019}),
  \bibinfo{pages}{101533}.
\newblock
\showISSN{22106707}
\urldef\tempurl%
\url{https://doi.org/10.1016/j.scs.2019.101533}
\showDOI{\tempurl}


\bibitem[Cao et~al\mbox{.}(2022)]%
        {cao_probabilistic_2022}
\bibfield{author}{\bibinfo{person}{Yang Cao}, \bibinfo{person}{Zhenzhen Dang},
  \bibinfo{person}{Feng Wu}, \bibinfo{person}{Xovee Xu}, {and}
  \bibinfo{person}{Fan Zhou}.} \bibinfo{year}{2022}\natexlab{}.
\newblock \showarticletitle{Probabilistic {Electricity} {Demand} {Forecasting}
  with {Transformer}-{Guided} {State} {Space} {Model}}. In
  \bibinfo{booktitle}{\emph{2022 {IEEE} 5th {International} {Conference} on
  {Automation}, {Electronics} and {Electrical} {Engineering} ({AUTEEE})}}.
  \bibinfo{pages}{964--969}.
\newblock
\urldef\tempurl%
\url{https://doi.org/10.1109/AUTEEE56487.2022.9994294}
\showDOI{\tempurl}


\bibitem[Edwards et~al\mbox{.}(2012)]%
        {edwards_predicting_2012}
\bibfield{author}{\bibinfo{person}{Richard~E. Edwards}, \bibinfo{person}{Joshua
  New}, {and} \bibinfo{person}{Lynne~E. Parker}.}
  \bibinfo{year}{2012}\natexlab{}.
\newblock \showarticletitle{Predicting future hourly residential electrical
  consumption: {A} machine learning case study}.
\newblock \bibinfo{journal}{\emph{Energy and Buildings}}  \bibinfo{volume}{49}
  (\bibinfo{date}{June} \bibinfo{year}{2012}), \bibinfo{pages}{591--603}.
\newblock
\showISSN{03787788}
\urldef\tempurl%
\url{https://doi.org/10.1016/j.enbuild.2012.03.010}
\showDOI{\tempurl}


\bibitem[Farsi et~al\mbox{.}(2021)]%
        {farsi_short-term_2021}
\bibfield{author}{\bibinfo{person}{Behnam Farsi}, \bibinfo{person}{Manar
  Amayri}, \bibinfo{person}{Nizar Bouguila}, {and} \bibinfo{person}{Ursula
  Eicker}.} \bibinfo{year}{2021}\natexlab{}.
\newblock \showarticletitle{On {Short}-{Term} {Load} {Forecasting} {Using}
  {Machine} {Learning} {Techniques} and a {Novel} {Parallel} {Deep}
  {LSTM}-{CNN} {Approach}}.
\newblock \bibinfo{journal}{\emph{IEEE Access}}  \bibinfo{volume}{9}
  (\bibinfo{year}{2021}), \bibinfo{pages}{31191--31212}.
\newblock
\showISSN{2169-3536}
\urldef\tempurl%
\url{https://doi.org/10.1109/ACCESS.2021.3060290}
\showDOI{\tempurl}


\bibitem[Groß et~al\mbox{.}(2021)]%
        {gros_comparison_2021}
\bibfield{author}{\bibinfo{person}{Arne Groß}, \bibinfo{person}{Antonia
  Lenders}, \bibinfo{person}{Friedhelm Schwenker}, \bibinfo{person}{Daniel~A.
  Braun}, {and} \bibinfo{person}{David Fischer}.}
  \bibinfo{year}{2021}\natexlab{}.
\newblock \showarticletitle{Comparison of short-term electrical load
  forecasting methods for different building types}.
\newblock \bibinfo{journal}{\emph{Energy Informatics}} \bibinfo{volume}{4},
  \bibinfo{number}{3} (\bibinfo{date}{Sept.} \bibinfo{year}{2021}),
  \bibinfo{pages}{13}.
\newblock
\showISSN{2520-8942}
\urldef\tempurl%
\url{https://doi.org/10.1186/s42162-021-00172-6}
\showDOI{\tempurl}


\bibitem[Haag et~al\mbox{.}(2022)]%
        {haag_augmented_2022}
\bibfield{author}{\bibinfo{person}{Felix Haag}, \bibinfo{person}{Konstantin
  Hopf}, \bibinfo{person}{Pedro~Menelau Vasconcelos}, {and}
  \bibinfo{person}{Thorsten Staake}.} \bibinfo{year}{2022}\natexlab{}.
\newblock \showarticletitle{Augmented {Cross}-{Selling} {Through} {Explainable}
  {AI}—{A} {Case} {From} {Energy} {Retailing}}. In
  \bibinfo{booktitle}{\emph{{ECIS} 2022 {Research} {Papers}}}.
  \bibinfo{publisher}{AIS electronic library}, \bibinfo{address}{Timisoara,
  Romania}.
\newblock
\urldef\tempurl%
\url{https://aisel.aisnet.org/ecis2022_rp/129}
\showURL{%
\tempurl}


\bibitem[Haben et~al\mbox{.}(2019)]%
        {haben_short_2019}
\bibfield{author}{\bibinfo{person}{Stephen Haben}, \bibinfo{person}{Georgios
  Giasemidis}, \bibinfo{person}{Florian Ziel}, {and} \bibinfo{person}{Siddharth
  Arora}.} \bibinfo{year}{2019}\natexlab{}.
\newblock \showarticletitle{Short {Term} {Load} {Forecasts} of {Low} {Voltage}
  {Demand} and the {Effects} of {Weather}}.
\newblock \bibinfo{journal}{\emph{International Journal of Forecasting}}
  \bibinfo{volume}{35}, \bibinfo{number}{4} (\bibinfo{date}{Oct.}
  \bibinfo{year}{2019}), \bibinfo{pages}{1469--1484}.
\newblock
\showISSN{01692070}
\urldef\tempurl%
\url{https://doi.org/10.1016/j.ijforecast.2018.10.007}
\showDOI{\tempurl}
\newblock
\shownote{arXiv: 1804.02955}.


\bibitem[Heinemann et~al\mbox{.}(1966)]%
        {heinemann_and_1966}
\bibfield{author}{\bibinfo{person}{G~T Heinemann}, \bibinfo{person}{D~A
  Nordmian}, {and} \bibinfo{person}{E~C Plant}.}
  \bibinfo{year}{1966}\natexlab{}.
\newblock \showarticletitle{and {Summer} {Loads}-{A} {Regression} {Analysis}}.
\newblock \bibinfo{journal}{\emph{IEEE TRANSACTIONS ON POWER APPARATUS AND
  SYSTEMS}} (\bibinfo{year}{1966}), \bibinfo{pages}{11}.
\newblock


\bibitem[Herzen et~al\mbox{.}(2022)]%
        {herzen_darts_2022}
\bibfield{author}{\bibinfo{person}{Julien Herzen}, \bibinfo{person}{Francesco
  Lässig}, \bibinfo{person}{Samuele~Giuliano Piazzetta},
  \bibinfo{person}{Thomas Neuer}, \bibinfo{person}{Léo Tafti},
  \bibinfo{person}{Guillaume Raille}, \bibinfo{person}{Tomas~Van Pottelbergh},
  \bibinfo{person}{Marek Pasieka}, \bibinfo{person}{Andrzej Skrodzki},
  \bibinfo{person}{Nicolas Huguenin}, \bibinfo{person}{Maxime Dumonal},
  \bibinfo{person}{Jan Kościsz}, \bibinfo{person}{Dennis Bader},
  \bibinfo{person}{Frédérick Gusset}, \bibinfo{person}{Mounir Benheddi},
  \bibinfo{person}{Camila Williamson}, \bibinfo{person}{Michal Kosinski},
  \bibinfo{person}{Matej Petrik}, {and} \bibinfo{person}{Gaël Grosch}.}
  \bibinfo{year}{2022}\natexlab{}.
\newblock \showarticletitle{Darts: {User}-{Friendly} {Modern} {Machine}
  {Learning} for {Time} {Series}}.
\newblock \bibinfo{journal}{\emph{Journal of Machine Learning Research}}
  \bibinfo{volume}{23}, \bibinfo{number}{124} (\bibinfo{year}{2022}),
  \bibinfo{pages}{1--6}.
\newblock
\urldef\tempurl%
\url{http://jmlr.org/papers/v23/21-1177.html}
\showURL{%
\tempurl}


\bibitem[Hong and Fan(2016)]%
        {hong_probabilistic_2016}
\bibfield{author}{\bibinfo{person}{Tao Hong} {and} \bibinfo{person}{Shu Fan}.}
  \bibinfo{year}{2016}\natexlab{}.
\newblock \showarticletitle{Probabilistic electric load forecasting: {A}
  tutorial review}.
\newblock \bibinfo{journal}{\emph{International Journal of Forecasting}}
  \bibinfo{volume}{32}, \bibinfo{number}{3} (\bibinfo{date}{July}
  \bibinfo{year}{2016}), \bibinfo{pages}{914--938}.
\newblock
\showISSN{01692070}
\urldef\tempurl%
\url{https://doi.org/10.1016/j.ijforecast.2015.11.011}
\showDOI{\tempurl}


\bibitem[Hong et~al\mbox{.}(2014)]%
        {hong_global_2014}
\bibfield{author}{\bibinfo{person}{Tao Hong}, \bibinfo{person}{Pierre Pinson},
  {and} \bibinfo{person}{Shu Fan}.} \bibinfo{year}{2014}\natexlab{}.
\newblock \showarticletitle{Global {Energy} {Forecasting} {Competition} 2012}.
\newblock \bibinfo{journal}{\emph{International Journal of Forecasting}}
  \bibinfo{volume}{30}, \bibinfo{number}{2} (\bibinfo{date}{April}
  \bibinfo{year}{2014}), \bibinfo{pages}{357--363}.
\newblock
\showISSN{01692070}
\urldef\tempurl%
\url{https://doi.org/10.1016/j.ijforecast.2013.07.001}
\showDOI{\tempurl}


\bibitem[Hong et~al\mbox{.}(2020)]%
        {hong_energy_2020}
\bibfield{author}{\bibinfo{person}{Tao Hong}, \bibinfo{person}{Pierre Pinson},
  \bibinfo{person}{Yi Wang}, \bibinfo{person}{Rafał Weron},
  \bibinfo{person}{Dazhi Yang}, {and} \bibinfo{person}{Hamidreza Zareipour}.}
  \bibinfo{year}{2020}\natexlab{}.
\newblock \showarticletitle{Energy {Forecasting}: {A} {Review} and {Outlook}}.
\newblock \bibinfo{journal}{\emph{IEEE Open Access Journal of Power and
  Energy}}  \bibinfo{volume}{7} (\bibinfo{date}{Oct.} \bibinfo{year}{2020}).
\newblock
\urldef\tempurl%
\url{https://doi.org/10.1109/OAJPE.2020.3029979}
\showDOI{\tempurl}


\bibitem[Hopf(2019)]%
        {hopf_predictive_2019}
\bibfield{author}{\bibinfo{person}{Konstantin Hopf}.}
  \bibinfo{year}{2019}\natexlab{}.
\newblock \bibinfo{booktitle}{\emph{Predictive {Analytics} for {Energy}
  {Efficiency} and {Energy} {Retailing}} (\bibinfo{edition}{1} ed.)}.
  \bibinfo{series}{Contributions of the {Faculty} {Information} {Systems} and
  {Applied} {Computer} {Sciences} of the {Otto}-{Friedrich}-{University}
  {Bamberg}}, Vol.~\bibinfo{volume}{36}.
\newblock \bibinfo{publisher}{University of Bamberg},
  \bibinfo{address}{Bamberg}.
\newblock
\showISBNx{978-3-86309-669-4}
\urldef\tempurl%
\url{https://doi.org/10.20378/irbo-54833}
\showDOI{\tempurl}


\bibitem[Huang et~al\mbox{.}(2022)]%
        {huang_short-term_2022}
\bibfield{author}{\bibinfo{person}{Shichao Huang}, \bibinfo{person}{Jing
  Zhang}, \bibinfo{person}{Yu He}, \bibinfo{person}{Xiaofan Fu},
  \bibinfo{person}{Luqin Fan}, \bibinfo{person}{Gang Yao}, {and}
  \bibinfo{person}{Yongjun Wen}.} \bibinfo{year}{2022}\natexlab{}.
\newblock \showarticletitle{Short-{Term} {Load} {Forecasting} {Based} on the
  {CEEMDAN}-{Sample} {Entropy}-{BPNN}-{Transformer}}.
\newblock \bibinfo{journal}{\emph{Energies}} \bibinfo{volume}{15},
  \bibinfo{number}{10} (\bibinfo{date}{Jan.} \bibinfo{year}{2022}),
  \bibinfo{pages}{3659}.
\newblock
\showISSN{1996-1073}
\urldef\tempurl%
\url{https://doi.org/10.3390/en15103659}
\showDOI{\tempurl}


\bibitem[Huy et~al\mbox{.}(2022)]%
        {huy_short-term_2022}
\bibfield{author}{\bibinfo{person}{Pham~Canh Huy}, \bibinfo{person}{Nguyen~Quoc
  Minh}, \bibinfo{person}{Nguyen~Dang Tien}, {and} \bibinfo{person}{Tao
  Thi~Quynh Anh}.} \bibinfo{year}{2022}\natexlab{}.
\newblock \showarticletitle{Short-{Term} {Electricity} {Load} {Forecasting}
  {Based} on {Temporal} {Fusion} {Transformer} {Model}}.
\newblock \bibinfo{journal}{\emph{IEEE Access}}  \bibinfo{volume}{10}
  (\bibinfo{year}{2022}), \bibinfo{pages}{106296--106304}.
\newblock
\showISSN{2169-3536}
\urldef\tempurl%
\url{https://doi.org/10.1109/ACCESS.2022.3211941}
\showDOI{\tempurl}


\bibitem[Høverstad et~al\mbox{.}(2015)]%
        {hoverstad_short-term_2015}
\bibfield{author}{\bibinfo{person}{Boye~A. Høverstad}, \bibinfo{person}{Axel
  Tidemann}, \bibinfo{person}{Helge Langseth}, {and} \bibinfo{person}{Pinar
  Öztürk}.} \bibinfo{year}{2015}\natexlab{}.
\newblock \showarticletitle{Short-{Term} {Load} {Forecasting} {With} {Seasonal}
  {Decomposition} {Using} {Evolution} for {Parameter} {Tuning}}.
\newblock \bibinfo{journal}{\emph{IEEE Transactions on Smart Grid}}
  \bibinfo{volume}{6}, \bibinfo{number}{4} (\bibinfo{date}{July}
  \bibinfo{year}{2015}), \bibinfo{pages}{1904--1913}.
\newblock
\showISSN{1949-3061}
\urldef\tempurl%
\url{https://doi.org/10.1109/TSG.2015.2395822}
\showDOI{\tempurl}


\bibitem[Jain et~al\mbox{.}(2014)]%
        {jain_forecasting_2014}
\bibfield{author}{\bibinfo{person}{Rishee~K. Jain}, \bibinfo{person}{Kevin~M.
  Smith}, \bibinfo{person}{Patricia~J. Culligan}, {and}
  \bibinfo{person}{John~E. Taylor}.} \bibinfo{year}{2014}\natexlab{}.
\newblock \showarticletitle{Forecasting energy consumption of multi-family
  residential buildings using support vector regression: {Investigating} the
  impact of temporal and spatial monitoring granularity on performance
  accuracy}.
\newblock \bibinfo{journal}{\emph{Applied Energy}}  \bibinfo{volume}{123}
  (\bibinfo{date}{June} \bibinfo{year}{2014}), \bibinfo{pages}{168--178}.
\newblock
\showISSN{03062619}
\urldef\tempurl%
\url{https://doi.org/10.1016/j.apenergy.2014.02.057}
\showDOI{\tempurl}


\bibitem[Khan et~al\mbox{.}(2015)]%
        {khan2015load}
\bibfield{author}{\bibinfo{person}{Ahsan~Raza Khan}, \bibinfo{person}{Anzar
  Mahmood}, \bibinfo{person}{Awais Safdar}, \bibinfo{person}{Zafar~A Khan},
  \bibinfo{person}{Syed Bilal}, {and} \bibinfo{person}{Naveed Ahmed~Khan
  Javaid}.} \bibinfo{year}{2015}\natexlab{}.
\newblock \showarticletitle{Load Forecasting and Dynamic Pricing based Energy
  Management in Smart Grid-A Review}. In
  \bibinfo{booktitle}{\emph{International Multi-topic Conference}}.
\newblock


\bibitem[Lago et~al\mbox{.}(2018)]%
        {lago_forecasting_2018}
\bibfield{author}{\bibinfo{person}{Jesus Lago}, \bibinfo{person}{Fjo
  De~Ridder}, {and} \bibinfo{person}{Bart De~Schutter}.}
  \bibinfo{year}{2018}\natexlab{}.
\newblock \showarticletitle{Forecasting spot electricity prices: {Deep}
  learning approaches and empirical comparison of traditional algorithms}.
\newblock \bibinfo{journal}{\emph{Applied Energy}}  \bibinfo{volume}{221}
  (\bibinfo{date}{July} \bibinfo{year}{2018}), \bibinfo{pages}{386--405}.
\newblock
\showISSN{03062619}
\urldef\tempurl%
\url{https://doi.org/10.1016/j.apenergy.2018.02.069}
\showDOI{\tempurl}


\bibitem[Li et~al\mbox{.}(2023)]%
        {li_probabilistic_2023}
\bibfield{author}{\bibinfo{person}{Dan Li}, \bibinfo{person}{Ya Tan},
  \bibinfo{person}{Yuanhang Zhang}, \bibinfo{person}{Shuwei Miao}, {and}
  \bibinfo{person}{Shuai He}.} \bibinfo{year}{2023}\natexlab{}.
\newblock \showarticletitle{Probabilistic forecasting method for mid-term
  hourly load time series based on an improved temporal fusion transformer
  model}.
\newblock \bibinfo{journal}{\emph{International Journal of Electrical Power \&
  Energy Systems}}  \bibinfo{volume}{146} (\bibinfo{date}{March}
  \bibinfo{year}{2023}), \bibinfo{pages}{108743}.
\newblock
\showISSN{0142-0615}
\urldef\tempurl%
\url{https://doi.org/10.1016/j.ijepes.2022.108743}
\showDOI{\tempurl}


\bibitem[Li et~al\mbox{.}(2022)]%
        {li_heat_2022}
\bibfield{author}{\bibinfo{person}{Guangxia Li}, \bibinfo{person}{Cheng Zhou},
  \bibinfo{person}{Ruiyu Li}, {and} \bibinfo{person}{Jia Liu}.}
  \bibinfo{year}{2022}\natexlab{}.
\newblock \showarticletitle{Heat load forecasting for district water-heating
  system using locality-enhanced transformer encoder}. In
  \bibinfo{booktitle}{\emph{Proceedings of the {Thirteenth} {ACM}
  {International} {Conference} on {Future} {Energy} {Systems}}}
  \emph{(\bibinfo{series}{e-{Energy} '22})}. \bibinfo{publisher}{Association
  for Computing Machinery}, \bibinfo{address}{New York, NY, USA},
  \bibinfo{pages}{440--441}.
\newblock
\showISBNx{978-1-4503-9397-3}
\urldef\tempurl%
\url{https://doi.org/10.1145/3538637.3538751}
\showDOI{\tempurl}


\bibitem[Lim et~al\mbox{.}(2021a)]%
        {lim_temporal_2020}
\bibfield{author}{\bibinfo{person}{Bryan Lim}, \bibinfo{person}{Sercan~Ö.
  Arık}, \bibinfo{person}{Nicolas Loeff}, {and} \bibinfo{person}{Tomas
  Pfister}.} \bibinfo{year}{2021}\natexlab{a}.
\newblock \showarticletitle{Temporal Fusion Transformers for interpretable
  multi-horizon time series forecasting}.
\newblock \bibinfo{journal}{\emph{International Journal of Forecasting}}
  \bibinfo{volume}{37}, \bibinfo{number}{4} (\bibinfo{year}{2021}),
  \bibinfo{pages}{1748--1764}.
\newblock
\showISSN{0169-2070}
\urldef\tempurl%
\url{https://doi.org/10.1016/j.ijforecast.2021.03.012}
\showDOI{\tempurl}


\bibitem[Lim et~al\mbox{.}(2021b)]%
        {lim_temporal_2021}
\bibfield{author}{\bibinfo{person}{Bryan Lim}, \bibinfo{person}{Sercan~Ö.
  Arık}, \bibinfo{person}{Nicolas Loeff}, {and} \bibinfo{person}{Tomas
  Pfister}.} \bibinfo{year}{2021}\natexlab{b}.
\newblock \showarticletitle{Temporal {Fusion} {Transformers} for interpretable
  multi-horizon time series forecasting}.
\newblock \bibinfo{journal}{\emph{International Journal of Forecasting}}
  \bibinfo{volume}{37}, \bibinfo{number}{4} (\bibinfo{date}{Oct.}
  \bibinfo{year}{2021}), \bibinfo{pages}{1748--1764}.
\newblock
\showISSN{0169-2070}
\urldef\tempurl%
\url{https://doi.org/10.1016/j.ijforecast.2021.03.012}
\showDOI{\tempurl}


\bibitem[L’Heureux et~al\mbox{.}(2022)]%
        {lheureux_transformer-based_2022}
\bibfield{author}{\bibinfo{person}{Alexandra L’Heureux},
  \bibinfo{person}{Katarina Grolinger}, {and} \bibinfo{person}{Miriam A.~M.
  Capretz}.} \bibinfo{year}{2022}\natexlab{}.
\newblock \showarticletitle{Transformer-{Based} {Model} for {Electrical} {Load}
  {Forecasting}}.
\newblock \bibinfo{journal}{\emph{Energies}} \bibinfo{volume}{15},
  \bibinfo{number}{14} (\bibinfo{date}{Jan.} \bibinfo{year}{2022}),
  \bibinfo{pages}{4993}.
\newblock
\showISSN{1996-1073}
\urldef\tempurl%
\url{https://doi.org/10.3390/en15144993}
\showDOI{\tempurl}


\bibitem[Mujeeb et~al\mbox{.}(2019)]%
        {mujeeb_deep_2019}
\bibfield{author}{\bibinfo{person}{Sana Mujeeb}, \bibinfo{person}{Nadeem
  Javaid}, \bibinfo{person}{Manzoor Ilahi}, \bibinfo{person}{Zahid Wadud},
  \bibinfo{person}{Farruh Ishmanov}, {and} \bibinfo{person}{Muhammad Afzal}.}
  \bibinfo{year}{2019}\natexlab{}.
\newblock \showarticletitle{Deep Long Short-Term Memory: A New Price and Load
  Forecasting Scheme for Big Data in Smart Cities}.
\newblock \bibinfo{journal}{\emph{Sustainability}}  \bibinfo{volume}{11}
  (\bibinfo{date}{02} \bibinfo{year}{2019}), \bibinfo{pages}{987}.
\newblock
\urldef\tempurl%
\url{https://doi.org/10.3390/su11040987}
\showDOI{\tempurl}


\bibitem[Shohan et~al\mbox{.}(2022)]%
        {shohan_forecasting_2022}
\bibfield{author}{\bibinfo{person}{Md~Jamal~Ahmed Shohan},
  \bibinfo{person}{Md~Omar Faruque}, {and} \bibinfo{person}{Simon~Y. Foo}.}
  \bibinfo{year}{2022}\natexlab{}.
\newblock \showarticletitle{Forecasting of {Electric} {Load} {Using} a {Hybrid}
  {LSTM}-{Neural} {Prophet} {Model}}.
\newblock \bibinfo{journal}{\emph{Energies}} \bibinfo{volume}{15},
  \bibinfo{number}{6} (\bibinfo{date}{March} \bibinfo{year}{2022}),
  \bibinfo{pages}{2158}.
\newblock
\showISSN{1996-1073}
\urldef\tempurl%
\url{https://doi.org/10.3390/en15062158}
\showDOI{\tempurl}


\bibitem[Sola and Sevilla(1997)]%
        {normalization}
\bibfield{author}{\bibinfo{person}{J. Sola} {and} \bibinfo{person}{J.
  Sevilla}.} \bibinfo{year}{1997}\natexlab{}.
\newblock \showarticletitle{Importance of input data normalization for the
  application of neural networks to complex industrial problems}.
\newblock \bibinfo{journal}{\emph{IEEE Transactions on Nuclear Science}}
  \bibinfo{volume}{44}, \bibinfo{number}{3} (\bibinfo{year}{1997}),
  \bibinfo{pages}{1464--1468}.
\newblock
\urldef\tempurl%
\url{https://doi.org/10.1109/23.589532}
\showDOI{\tempurl}


\bibitem[Vaswani et~al\mbox{.}(2017)]%
        {vaswani_attention_2017}
\bibfield{author}{\bibinfo{person}{Ashish Vaswani}, \bibinfo{person}{Noam
  Shazeer}, \bibinfo{person}{Niki Parmar}, \bibinfo{person}{Jakob Uszkoreit},
  \bibinfo{person}{Llion Jones}, \bibinfo{person}{Aidan~N. Gomez},
  \bibinfo{person}{Lukasz Kaiser}, {and} \bibinfo{person}{Illia Polosukhin}.}
  \bibinfo{year}{2017}\natexlab{}.
\newblock \showarticletitle{Attention {Is} {All} {You} {Need}}.
\newblock \bibinfo{journal}{\emph{arXiv:1706.03762 [cs]}} (\bibinfo{date}{Dec.}
  \bibinfo{year}{2017}).
\newblock
\urldef\tempurl%
\url{http://arxiv.org/abs/1706.03762}
\showURL{%
\tempurl}
\newblock
\shownote{arXiv: 1706.03762}.


\bibitem[vom Scheidt et~al\mbox{.}(2020)]%
        {vom_scheidt_data_2020}
\bibfield{author}{\bibinfo{person}{Frederik vom Scheidt}, \bibinfo{person}{Hana
  Medinová}, \bibinfo{person}{Nicole Ludwig}, \bibinfo{person}{Bent Richter},
  \bibinfo{person}{Philipp Staudt}, {and} \bibinfo{person}{Christof
  Weinhardt}.} \bibinfo{year}{2020}\natexlab{}.
\newblock \showarticletitle{Data analytics in the electricity sector – {A}
  quantitative and qualitative literature review}.
\newblock \bibinfo{journal}{\emph{Energy and AI}}  \bibinfo{volume}{1}
  (\bibinfo{year}{2020}), \bibinfo{pages}{Article no: 100009}.
\newblock
\showISSN{2666-5468}
\urldef\tempurl%
\url{https://doi.org/10.1016/j.egyai.2020.100009}
\showDOI{\tempurl}


\bibitem[Wang et~al\mbox{.}(2019)]%
        {wang_review_2019}
\bibfield{author}{\bibinfo{person}{Yi Wang}, \bibinfo{person}{Qixin Chen},
  \bibinfo{person}{Tao Hong}, {and} \bibinfo{person}{Chongqing Kang}.}
  \bibinfo{year}{2019}\natexlab{}.
\newblock \showarticletitle{Review of {Smart} {Meter} {Data} {Analytics}:
  {Applications}, {Methodologies}, and {Challenges}}.
\newblock \bibinfo{journal}{\emph{IEEE Transactions on Smart Grid}}
  \bibinfo{volume}{10}, \bibinfo{number}{3} (\bibinfo{date}{May}
  \bibinfo{year}{2019}), \bibinfo{pages}{3125--3148}.
\newblock
\showISSN{1949-3061}
\urldef\tempurl%
\url{https://doi.org/10.1109/TSG.2018.2818167}
\showDOI{\tempurl}
\newblock
\shownote{Conference Name: IEEE Transactions on Smart Grid}.


\bibitem[Wen et~al\mbox{.}(2022)]%
        {wen_impact_2022}
\bibfield{author}{\bibinfo{person}{Le Wen}, \bibinfo{person}{Basil Sharp},
  \bibinfo{person}{Kiti Suomalainen}, \bibinfo{person}{Mingyue~Selena Sheng},
  {and} \bibinfo{person}{Fengtao Guang}.} \bibinfo{year}{2022}\natexlab{}.
\newblock \showarticletitle{The impact of {COVID}-19 containment measures on
  changes in electricity demand}.
\newblock \bibinfo{journal}{\emph{Sustainable Energy, Grids and Networks}}
  \bibinfo{volume}{29} (\bibinfo{date}{March} \bibinfo{year}{2022}),
  \bibinfo{pages}{100571}.
\newblock
\showISSN{2352-4677}
\urldef\tempurl%
\url{https://doi.org/10.1016/j.segan.2021.100571}
\showDOI{\tempurl}


\bibitem[Zhang et~al\mbox{.}(2022)]%
        {zhang_short-term_2022}
\bibfield{author}{\bibinfo{person}{Guangqi Zhang}, \bibinfo{person}{Chuyuan
  Wei}, \bibinfo{person}{Changfeng Jing}, {and} \bibinfo{person}{Yanxue Wang}.}
  \bibinfo{year}{2022}\natexlab{}.
\newblock \showarticletitle{Short-{Term} {Electrical} {Load} {Forecasting}
  {Based} on {Time} {Augmented} {Transformer}}.
\newblock \bibinfo{journal}{\emph{International Journal of Computational
  Intelligence Systems}} \bibinfo{volume}{15}, \bibinfo{number}{1}
  (\bibinfo{date}{Aug.} \bibinfo{year}{2022}), \bibinfo{pages}{67}.
\newblock
\showISSN{1875-6883}
\urldef\tempurl%
\url{https://doi.org/10.1007/s44196-022-00128-y}
\showDOI{\tempurl}


\bibitem[Zhang et~al\mbox{.}(2021)]%
        {zhang_power_2021}
\bibfield{author}{\bibinfo{person}{Junfeng Zhang}, \bibinfo{person}{Hui Zhang},
  \bibinfo{person}{Song Ding}, {and} \bibinfo{person}{Xiaoxiong Zhang}.}
  \bibinfo{year}{2021}\natexlab{}.
\newblock \showarticletitle{Power {Consumption} {Predicting} and {Anomaly}
  {Detection} {Based} on {Transformer} and {K}-{Means}}.
\newblock \bibinfo{journal}{\emph{Frontiers in Energy Research}}
  \bibinfo{volume}{9} (\bibinfo{date}{Oct.} \bibinfo{year}{2021}),
  \bibinfo{pages}{779587}.
\newblock
\showISSN{2296-598X}
\urldef\tempurl%
\url{https://doi.org/10.3389/fenrg.2021.779587}
\showDOI{\tempurl}


\bibitem[Zhao et~al\mbox{.}(2021)]%
        {zhao_short-term_2021}
\bibfield{author}{\bibinfo{person}{Zezheng Zhao}, \bibinfo{person}{Chunqiu
  Xia}, \bibinfo{person}{Lian Chi}, \bibinfo{person}{Xiaomin Chang},
  \bibinfo{person}{Wei Li}, \bibinfo{person}{Ting Yang}, {and}
  \bibinfo{person}{Albert~Y. Zomaya}.} \bibinfo{year}{2021}\natexlab{}.
\newblock \showarticletitle{Short-{Term} {Load} {Forecasting} {Based} on the
  {Transformer} {Model}}.
\newblock \bibinfo{journal}{\emph{Information}} \bibinfo{volume}{12},
  \bibinfo{number}{12} (\bibinfo{date}{Dec.} \bibinfo{year}{2021}),
  \bibinfo{pages}{516}.
\newblock
\showISSN{2078-2489}
\urldef\tempurl%
\url{https://doi.org/10.3390/info12120516}
\showDOI{\tempurl}


\bibitem[Zheng et~al\mbox{.}(2017)]%
        {zheng_short_2017}
\bibfield{author}{\bibinfo{person}{Huiting Zheng}, \bibinfo{person}{Jiabin
  Yuan}, {and} \bibinfo{person}{Long Chen}.} \bibinfo{year}{2017}\natexlab{}.
\newblock \showarticletitle{Short-term load forecasting using EMD-LSTM neural
  networks with a Xgboost algorithm for feature importance evaluation}.
\newblock \bibinfo{journal}{\emph{Energies}} \bibinfo{volume}{10},
  \bibinfo{number}{8} (\bibinfo{year}{2017}), \bibinfo{pages}{1168}.
\newblock


\bibitem[Zhong et~al\mbox{.}(2020)]%
        {zhong_implications_2020}
\bibfield{author}{\bibinfo{person}{Haiwang Zhong}, \bibinfo{person}{Zhenfei
  Tan}, \bibinfo{person}{Yiliu He}, \bibinfo{person}{Le Xie}, {and}
  \bibinfo{person}{Chongqing Kang}.} \bibinfo{year}{2020}\natexlab{}.
\newblock \showarticletitle{Implications of {COVID}-19 for the electricity
  industry: {A} comprehensive review}.
\newblock \bibinfo{journal}{\emph{CSEE Journal of Power and Energy Systems}}
  \bibinfo{volume}{6}, \bibinfo{number}{3} (\bibinfo{date}{Sept.}
  \bibinfo{year}{2020}), \bibinfo{pages}{489--495}.
\newblock
\showISSN{2096-0042}
\urldef\tempurl%
\url{https://doi.org/10.17775/CSEEJPES.2020.02500}
\showDOI{\tempurl}


\bibitem[Zou et~al\mbox{.}(2019)]%
        {zou_weather_2019}
\bibfield{author}{\bibinfo{person}{Mingzhe Zou}, \bibinfo{person}{Duo Fang},
  \bibinfo{person}{Gareth Harrison}, {and} \bibinfo{person}{Sasa Djokic}.}
  \bibinfo{year}{2019}\natexlab{}.
\newblock \showarticletitle{Weather {Based} {Day}-{Ahead} and {Week}-{Ahead}
  {Load} {Forecasting} using {Deep} {Recurrent} {Neural} {Network}}. In
  \bibinfo{booktitle}{\emph{2019 {IEEE} 5th {International} forum on {Research}
  and {Technology} for {Society} and {Industry} ({RTSI})}}.
  \bibinfo{publisher}{IEEE}, \bibinfo{address}{Florence, Italy},
  \bibinfo{pages}{341--346}.
\newblock
\showISBNx{978-1-72813-815-2}
\urldef\tempurl%
\url{https://doi.org/10.1109/RTSI.2019.8895580}
\showDOI{\tempurl}


\end{thebibliography}

\clearpage
\appendix
\section{Data Processing}
\subsection{Datasets}
The two real datasets that we select stem from two geographic regions, differ in the number of substations, the magnitude of load connected, and the timespans of the data. The severe differences in the datasets should ensure the external validity of our results. %For the development and testing of the proposed approach, we use two real datasets from distribution grids.  

\paragraph{US dataset} 
The dataset stems from the Global Energy Forecasting Competition 2012 (GEFC'12) \cite{hong_global_2014} and comprises data from 20 grid zones in the U.S., which we consider as substations. For our analyses, we consider the years 2004-2007 of the dataset. The authors of the dataset \cite{hong_global_2014} advise excluding two substations, \#4 because of outages and \#9 because it was covered by an industrial customer. We remove substation \#9 but keep substation \#4 and add the following data cleaning to all substations to handle potential outages on every substation.
We identify extreme values (e.g., outages) per substation using the statistical quartiles $q_{0.25}$ and $q_{0.75}$ and remove values smaller than $q_{0.25} - 1.5 * (q_{0.27}-q_{0.25})$, known as the inter quantile range. We calculate the quartiles per substation. Only demand values for substation \#4 fall under this criterion, and we remove $52$ out of $39{,}576$ (0.0013\%) data points. We replace the removed measurements using a linear interpolation \cite{gros_comparison_2021}. 

%The average consumption of the different substations (M=$20{,}681.34\akWh$, SD=$19{,}300.88\akWh$) ranges between $124.77\akWh$ and $53{,}274.63\akWh$.
%We aggregate the substations for the forecast on the grid-level which results in an average consumption of $392{,}945.48\akWh (\pm93{,}780.17$).

The dataset also contains temperature data of 11 weather stations in the U.S. but the  connection of the weather stations to the zones is not given (the connection of the weather stations to the grid zones was part of the GEFC'12 challenge). Hence, for our analysis, we use the average temperature of all 11 weather stations per hour. %This results in an average temperature of $14.36$ (SD=4.32) with minimum temperature 1.89 and maximum temperature 25.05.

%\paragraph{PT dataset} 
%To enhance replicability, we use a publicly available dataset from Portugal. We use the UCI Electricity Load Diagrams Dataset \cite{trindade_electricityloaddiagrams20112014_2015}, which contains the electricity consumption of 370 customers from the years 2011 until 2014. The raw data consists of 15-minute load data with readings in $kW$, which we transformed to $\akWh$ values on an hourly level, according to \cite{yu_temporal_2016, lim_temporal_2021}. The dataset's average load is $150{,}431.95\akWh$ (SD $153{,}567.20\akWh$). 
%lowest Value:  0.0
%highest Value:  764500.0
%Avg Value:  150431.95014830024
%Std Value:  153567.20302832098

\paragraph{DE dataset}
In addition to the public dataset, which does not contain detailed geographic references, we use a dataset from central Germany, which we obtained from a local distribution grid operator. The dataset consists of hourly smart meter data on the household level, covering the years 2019--2021 (36 months). Given that the first wave of the COVID-19 pandemic started in Germany in March 2020, the dataset contains 14 months of pre-pandemic electricity load and 22 months within the pandemic. In total, $9{,}455$ households are connected to one of 70 substations in the distribution grid, where each substation serves between 8 and 447 households (M=135.07, SD=94.75). 

We prepare the data on the level of each household and apply the following preparation steps. 
First, we remove households with unusually low consumption values. For this purpose, we exclude observations with a mean consumption less than $0.01$ kWh or a total consumption less than $100$ kWh. In total, we remove 598 households applying this criterion.
Second, we harmonize time shift events (to and from daylight saving time) in spring and fall. For time shifts in the fall, where a single day has 25 hours, we exclude the extra hour. For the days with only 23 hours (i.e., time shift in spring), we linearly interpolate the missing value to harmonize the data into a 24-hour shape. 
%
%Status Status Value Nr. of Occurrences
%true value w       20.267.386
%manual value m             21
%replacement value e    16.318
%provisional value v       170
%defective value g     116.943
%incorrect value f   1.229.854
%
%All values with the status variable in the \textit{remove-set} (status v, g and f, see Table \ref{tab:status-val}) were removed. 
Third, we remove all values from the meter readings that were labelled as "provisional", "defective", and "incorrect" and linearly interpolated the readings. In total, we replace $1{,}572{,}786$ of such missing values out of $51{,}157{,}455{,}768$ observations (0.0031\%).

Finally, we aggregate the data on the level of the substations and on the grid-level for our analysis. %The average consumption of all substations (M=$48.11$ kWh, SD=$27.52$ kWh) ranges between 1.68 kWh and 171.09 kWh.
%grid-levela
%hass_min cons:  1421.21
%hass max cons:  6026.36
%hass mean cons:  3175.82
%hass std cons:  881.27
We were also provided with the geographic location data for each substation, which we leverage to connect weather and epidemic data.

\subsection{Features}
\label{app:data_feats}
From the \textit{calendar data}, we extract the hour of the day, day of the week, day of the year, and a binary feature if a day is a national holiday or weekend day. We use the Python package \textit{python-holidays\footnote{\url{https://github.com/dr-prodigy/python-holidays}}} to obtain the local holidays. As most of the calendar features have a cyclic pattern, we encode them cyclically combining sine and cosine, following \cite{edwards_predicting_2012}. %
%The reason for that is for example, a week ends on a Sunday, but the following day is the new start of the week, the Monday. Neural networks work with numbers only. The day of the week, for example, is not handed to the network as a string, but the weekdays are integer encoded so that \textit{Monday} is $0$, \textit{Tuesday} is $1$ until \textit{Sunday} is encoded to a $6$.
%Thus, a method is required to express the underlying cyclic pattern the data has and allows neural networks to learn that the time distance between a Sunday and a Monday is not so much further than the time distance between Saturday and Sunday.

% Temperature
One of the most common input features in demand forecasts are meteorological variables, in particular, temperature data \cite{haben_short_2019, hoverstad_short-term_2015}. As \textit{weather data}, we use the hourly temperature of the region obtained from the \textit{Meteostat}\footnote{\url{https://meteostat.net/en/}} Python package, which uses, for example, data from the German Meteorological Service\footnote{\url{https://www.dwd.de/}}. For cities and areas with no own weather station, we interpolate the temperature for the selected geographical point using the geographic reference and altitude. We assume that the temperature data is also available for the test data horizon and apply the measurements as a proxy for a weather forecast.

Finally, we consider a data source that we have not found to be used by earlier studies, namely \textit{epidemic data}. This is feasible, as one of the datasets we include covers the beginning of the COVID-19 pandemic and thus accounts for multiple lockdowns in the area of the grid. This lead us to include the officially announced number of infected people in the area as a feature. Such data is, for example, published by the German Robert Koch Institute\footnote{\url{https://github.com/robert-koch-institut/SARS-CoV-2-Infektionen_in_Deutschland}}.
% 

%US Cons
%US Sub min:  43.75
%US Sub max:  135098.25
%US Sub mean:  20681.34
%US Sub std:  19300.88
%US grid min:  198189.25
%US grid min:  198189.25
%US grid mean:  392945.48
%US grid std:  93778.83
% temperature
%US Sub min temp:  1.89
%US Sub max temp:  25.05
%US Sub mean temp:  14.46
%US Sub std temp:  4.34
% US Special days: 458 von 1461 insgesamt

%DE Cons
%DE Sub min:  0.33
%DE Sub max:  359.9
%DE Sub mean:  46.93
%DE Sub std:  31.65
%DE grid min:  1421.21
%DE grid max:  6026.36
%DE grid mean:  3175.82
%DE grid std:  881.29
%DE Temperature
%DE grid min temp:  -16.5
%DE grid max temp:  37.8
%DE grid mean temp:  10.48
%DE grid std temp:  7.91
% DE special days 341 von 1096 Tagen insgesamt
%\section{Literature overview}

\section{Hyperparameters}
We list the value ranges for our hypoerparameter search in \autoref{tab:hyperparameter_ranges} and the finally used parameters in \autoref{tab:parameters}.
\begin{table}[h]
    %\small
    \caption{Value ranges hyperparameter tuning}
    \label{tab:hyperparameter_ranges}
    \begin{tabu}{XX}
        \toprule
        \multicolumn{2}{l}{TFT hyper-parameter}\\       
        \midrule
            att heads & [1, 4]\\
            hidden size & [16, 32, 64]	\\
            dropout& [0.1, 0.3]\\
            batch size & [32, 128]\\
            \ac{LSTM} layers & [1, 2, 4]\\
        \midrule
        \multicolumn{2}{l}{LSTM hyper-parameter}\\
        \midrule
        batch size & [50, 10, 120, 150]\\
        learning rate & [0.001, 0.01, 0.1]\\
        dropout & [0.1, 0.2, 0.3] \\
        \ac{LSTM} layer & [1, 2, 4]\\
        hidden size & [64, 128, 248, 496]	\\
        \bottomrule
    \end{tabu}
\end{table}

\begin{table*}[b]
    %\centering
    \footnotesize
    \caption{Final hyper-parameters}\label{tab:parameters}
    \begin{tabular}{p{1.8cm}p{1.5cm}p{0.8cm}p{0.8cm}p{0.8cm}p{0.8cm}p{0.8cm}p{0.8cm}p{0.8cm}p{0.8cm}p{0.8cm}p{0.8cm}p{0.8cm}p{0.8cm}}
    \toprule
     && \multicolumn{6}{c}{\textbf{TFT}} & \multicolumn{6}{c}{\textbf{LSTM }} \\
     \cmidrule(r){3-8}
     \cmidrule(r){9-14}
     Horizon & Dataset & att. heads & hidden size & \ac{LSTM} layers & input size & dropout & batch size & hidden size & num layers & input size & dropout & batch size & learning rate \\
     \midrule
     %\multicolumn{13}{l}{Grid-level forecast} \\
     %\midrule
     %Day $DE$ &&&&&&&&&&&&\\
     %Day $US$ &&&&&&&&&&&&\\
     %Week $DE$ &&&&&&&&&&&&\\
     %Week $US$ &&&&&&&&&&&&\\
     %\midrule
     \multicolumn{14}{l}{Grid-level forecast, consumption + calendar} \\
     \midrule
    Day & $DE$          &    1 &      64 &      2 &        24 &       0.1 &        32 &      64 &       2 &        48 &       0.2 &        10 &       0.01 \\
    Week &$DE$                &         1 &          64 &           4 &           504 &       0.1 &           32 &         64 &          4 &           336 &       0.1 &           10 &         0.001 \\
     Day & $US$                 &         1 &          64 &           2 &           336 &       0.3 &           32 &        496 &          1 &            48 &       0.2 &           10 &         0.001 \\
    Week & $US$               &         1 &          64 &           2 &           336 &       0.1 &           32 &        248 &          1 &           504 &       0.2 &           10 &         0.001 \\
    \midrule
    \multicolumn{14}{l}{Grid-level forecast, consumption + weather + calendar}\\
    \midrule
    Day & $DE$         &         4 &          64 &           2 &           672 &       0.3 &           32 &         64 &          4 &           168 &       0.1 &           10 &          0.01 \\
    Week & $DE$       &         4 &          64 &           4 &           672 &       0.1 &           32 &        128 &          1 &           504 &       0.3 &           10 &          0.01 \\
    Day & $US$        &         4 &          64 &           2 &           168 &       0.1 &           32 &        128 &          1 &           168 &       0.2 &           10 &          0.01 \\
    Week & $US$       &         1 &          64 &           4 &           168 &       0.1 &           32 &         64 &          1 &           672 &       0.2 &           10 &          0.01 \\
    \midrule
    \multicolumn{14}{l}{Grid-level, consumption + weather + calendar + epidemic features}\\
    \midrule
    Day & $DE$  &         4 &          32 &           1 &            48 &       0.1 &           32 &        x &        x &           x &       x &          x &           x \\
    Week & $DE$ &         4 &          64 &           2 &           168 &       0.1 &           32 &        x &        x &           x &       x &          x &           x \\
    \midrule
    \multicolumn{14}{l}{Hierarchical/substation forecast, consumption + weather + calendar}\\
    \midrule
    Day & $DE$         &         1 &          32 &           1 &           336 &       0.1 &          128 &        128 &          2 &            72 &       0.2 &          120 &          0.01 \\
    Week & $DE$        &         4 &          64 &           2 &           168 &       0.1 &          128 &        248 &          4 &           672 &       0.3 &           10 &         0.001 \\
    Day & $US$         &       4 &         64 &         2 &           72 &       0.1 &          31 &        128 &          2 &            48 &       0.1 &           10 &         0.001 \\
    Week & $US$        &       1 &         16 &         1 &           672 &       0.3 &          32 &         64 &          2 &           168 &       0.1 &           50 &         0.001 \\
    \midrule
    \multicolumn{14}{l}{Hierarchical/substation forecast, consumption + weather + calendar + epidemic feature}\\
    \midrule
    Day & $DE$   &         4 &          64 &           2 &           336 &       0.3 &           32 &        x &        x &           x &       x &          x &           x \\
    Week & $DE$  &         4 &          32 &           2 &           336 &       0.1 &           32 &        x &        x &           x &       x &          x &           x \\
\bottomrule
    \end{tabular}
    \end{table*}

\section{Detailed results}
See our detailed evaluation results in \autoref{tab:error-results}.

\begin{table*}[t]
    \footnotesize
    \centering
    \caption{Forecasting performance}\label{tab:error-results}
    \begin{tabu}{Xrrrrrrrrrrrr}
    \toprule
     & \multicolumn{6}{c}{\textbf{day-ahead}} & \multicolumn{6}{c}{\textbf{week-ahead }} \\
     \cmidrule(r){2-7}
     \cmidrule(r){8-13}
    \textbf{Model} & \multicolumn{2}{c}{\textbf{RMSE}} & \multicolumn{2}{c}{\textbf{MAPE}} & \multicolumn{2}{c}{\textbf{\ac{SMAPE}}} & \multicolumn{2}{c}{\textbf{RMSE}} & \multicolumn{2}{c}{\textbf{MAPE}} & \multicolumn{2}{c}{\textbf{\ac{SMAPE}}} \\ 
    \midrule
    \multicolumn{13}{l}{I. Grid-level forecast, demand only (both datasets)}\\
    \midrule
    \ac{ARIMA} $DE$ & 191{,}376.30 &($\pm 73{,}874.24$)  & 116.23   &($\pm 60.62$) & 183.82          &($\pm 22.69$) & 197{,}129.80    &($\pm 63{,}766.59$) & 115.25        &($\pm 42.65$) & 185.51     & ($\pm 18.69$)\\
    ARIMA $US$ &251{,}621.73 &($\pm 675{,}237.99$)&56.99 &($\pm161.73 $)&64.39	 &($\pm 14.66$)     &868{,}061.20 &($\pm2{,}431{,}013 $)&209.04 &($\pm 602.5 $)& 80.56 &($\pm 35.12 $)\\
    \midrule
    \multicolumn{13}{l}{II. Grid-level forecast, demand + calendar (both datasets)}\\
    \midrule
    \ac{LSTM} $DE$ &154.79        &($\pm68.68	$)         & 3.94         &($\pm1.52$)  & 4.01        &($\pm1.54$) & 190.73         &($\pm48.37$) & 4.94	       &($\pm0.91$) & 4.88         &($\pm0.84$)	\\
    \ac{TFT} $DE$ & 151.13        &($\pm58.52$)         & 3.98         &($\pm1.25$)   &  3.95           &($\pm1.22$)  & 167.12           &($\pm54.37$)         & 4.13         &($\pm0.92$)  &   4.18       &($\pm0.96$)\\
    \ac{LSTM} $US$ & 32{,}116.58        &($\pm16{,}425.95$)         & 6.25         &($\pm2.76$)  & 6.35        &($\pm2.88$)&50{,}107.86        &($\pm18{,}577.51$)&10.14         &($\pm4.27$)& 9.79         &($\pm3.67$)\\
    \ac{TFT} $US$ &30{,}977.48&($\pm 15{,}061.03$)&6.11 &($\pm 2.73$)&6.19 &($\pm 2.82$)     &50{,}232.94 &($\pm 17{,}623.17$)&9.98 &($\pm 3.23$)& 10.01 &($\pm 3.24$)\\
    %& $29{,}314.69   &(\pm12{,}901.99$)   & $11.49        &(\pm3.93$)   &  $11.08          &(\pm3.32$)  & $38{,}353.65      &(\pm18{,}281.83$)   & $14.38        &(\pm3.70$)  &  $13.89       &(\pm3.55$)\\
    \midrule
    % \multicolumn{13}{l}{Grid-level forecast, consumption only (transferability between the datasets)}\\
    % \midrule
    % \ac{TFT} $DE\rightarrow PT$  &?&&?&&?&&?&&?&&?\\ %&$101{,}106.89  &(\pm52{,}488.15$)   & $46.98        &(\pm15.58$)  &  $74.84          &(\pm72.54$) & $127{,}984.30     &(\pm52{,}138.87$)   & $64.85        &(\pm15.89$) &  $80.05       &(\pm67.49$) \\
    % \ac{TFT} $PT\rightarrow DE$&?&&?&&?&&?&&?&&?\\ %& $520.97        &(\pm176.65$)        & $12.60        &(\pm2.94$)   &  $13.73          &(\pm3.65$)  & $641.55           &(\pm276.00$)        & $16.20        &(\pm5.54$)  &  $18.03       &(\pm7.18$)\\
    % \midrule
    \multicolumn{13}{l}{III. Grid-level forecast, demand + weather + calendar}\\
    \midrule
    \ac{LSTM} $DE$              & 137.42& ($\pm 59.21$)     & 3.52 &($\pm 1.21$) & 3.54    &($\pm 1.23$) & 157.66          & ($\pm 39.60$)       & 4.24          &($\pm 1.13$) & 4.18         &($\pm 1.05$)\\
    \ac{TFT} $DE$               & 164.53        &($\pm 60.43$)       & 4.22          &($\pm 1.34$) & 4.20              &($\pm 1.31$) & 151.41           &($\pm 52.1$)        & 3.88 &($\pm 0.96$) & 3.83 &($\pm 0.9$) \\
    \ac{LSTM} $US$          & 25{,}531.28& ($\pm 14{,}523.90$)     & 4.89 &($\pm 2.35$) & 4.99    &($\pm 2.50$) & 57{,}549.48& ($\pm 24{,}345.73$) & 11.77 &($\pm 5.55$)& 11.26    &($\pm 4.99$) \\
    \ac{TFT} $US$          & 22{,}825.27& ($\pm 10{,}190.03$)     &4.59 &($\pm 1.86$) & 4.70    &($\pm1.98$) & 26{,}967.22& ($\pm 9{,}804.02$)     &5.22 &($\pm 1.80$) & 5.36 &($\pm 1.97$) \\
    \midrule
    
    \multicolumn{13}{l}{IV. Hierarchical forecast, demand + weather + calendar}\\
    \midrule
    \ac{LSTM} $DE$               & 146.43        &($\pm 67.89$)      & 3.65           &($\pm 1.42$) & 3.60              &($\pm 1.33$) & 337.97           &($\pm 119.81$)      & 7.99          &($\pm 1.98$) & 7.56 &($\pm 1.75$) \\
    \ac{TFT} $DE$                & 102.46        &($\pm 55.09$)      & 2.55           &($\pm 1.06$) & 2.54             &($\pm 1.05$) & 102.75           &($\pm 29.58$)       & \textbf{2.5}  &($\pm 0.49$) & 2.52 &($\pm 0.5$) \\
    \ac{LSTM} $US$                & 19{,}751.91        &($\pm 9{,}955.10	$)      & 3.88           &($\pm 2.02$) & 3.86             &($\pm 1.95$) & 32{,}064.27           &($\pm 10{,}200.25$)       & 6.23  &($\pm 1.88$) & 6.39 &($\pm 2.03	$) \\
    \ac{TFT} $US$                & 15{,}712.65        &($\pm 8{,}763.61$)      & 3.04           &($\pm 1.59$) & 3.09             &($\pm 1.65$) & 18{,}955.79           &($\pm 6{,}553.06$)       & 3.76  &($\pm 1.42$) & 3.81 &($\pm 1.49$) \\
    % \midrule
    % \multicolumn{13}{l}{Hierarchical forecast, consumption + weather + calendar (transferability between the datasets)}\\
    % \midrule
    % \ac{TFT} $DE\rightarrow US$              & 392324.04        &($\pm 59137.33$)      & $153{,}944{,}300$           &($\pm 72404475.38$) & 200              &($\pm 0$) &    1541899        &($\pm 184947.3$)      & 751834300          &($\pm 270423500$) & 200 &($\pm 0$) \\
    
    \midrule
    
    \multicolumn{13}{l}{V. Substation forecast, consumption + weather + calendar}\\
    \midrule
    \ac{LSTM} $DE$              & 7.26&($\pm 5.82$)      & 16.98 &($\pm 16.07$) & 28.70	    &($\pm 48.55$) & 10.97&($\pm8.73$)	 &27.58	&($\pm29.90$) &35.17 &($\pm47.92$)\\
    \ac{TFT} $DE$              &4.52 &($\pm 2.78$)      & 10.15	&($\pm 7.52$) & 23.38 & ($\pm 49.36$) &4.83 &($\pm2.71 $)  & 10.56& ($\pm7.65 $)& 23.91	&($\pm 49.29$)\\
    \ac{LSTM} $US$               & 1{,}518.69        &($\pm 1{,}592.79$)      & 6.54	           &($\pm 3.73$) & 6.45             &($\pm 3.50$) & 2{,}453.43           &($\pm 2{,}397.44$)       & 10.04  &($\pm 3.62$) & 10.07 &($\pm 3.57$)\\
    \ac{TFT} $US$                & 1{,}172.17        &($\pm 1{,}310.05$)      & 4.81           &($\pm 2.61$) & 4.85             &($\pm 2.62$) & 1{,}527.51           &($\pm 1{,}456.43$)       & 6.43  &($\pm 2.63$) & 6.38 &($\pm 2.51$) \\
    \midrule
    \multicolumn{13}{l}{VI. Forecast with demand + weather + calendar + epidemic features }\\
    \midrule
    \ac{TFT} $DE$ (grid-level)                & 169.59         &($\pm 63.46$)      & 4.46          &($\pm 1.2$)  & 4.48             &($\pm 1.2$)  & 149.39  &($\pm 44.35$)       & 3.89          &($\pm 0.61$) & 3.88         &($\pm 0.61$)\\
    %\midrule
    %\multicolumn{13}{l}{VII. Hierarchical forecast, demand + weather + calendar + epidemic features}\\
    %\midrule
    %\ac{LSTM}               &?&&?&&?&&?&&?&&? \\
    \ac{TFT} $DE$ (hierarchical)               & 98.32 &($\pm 50.49$)      & 2.43 &($\pm 0.88$) & 2.44    &($\pm 0.89$) &100.59   &($\pm 26.72$)       & 2.52          &($\pm 0.47$) & 2.51 &($\pm 0.46$)\\
    %\midrule
    %\multicolumn{13}{l}{VIII. Substation forecast, demand + weather + calendar + epidemic features }\\
    %\midrule
    \ac{TFT} $DE$ (substation)             &4.39&($\pm 2.81$)      & 9.86&($\pm 7.68$)      & 23.12&($\pm 49.42$)    & 4.84&($\pm 2.74$) & 10.85& ($\pm 8.05$)&23.99 &($\pm49.27 $)\\  
    \bottomrule
    \end{tabu}
\end{table*}

\end{document}